# Multi Scale Graph Wavenet for Wind Speed Forecasting

Neetesh Rathore[1], Pradeep Rathore[1], Arghya Basak[1], Sri Harsha Nistala, Venkataramana Runkana

TCS Research, Pune, 411013, India

Email: nrathore@ma.iitr.ac.in,{rathore.pradeep, arghya.basak, sriharsha.nistala, venkat.runkana}@tcs.com

*Abstract—* Geometric deep learning has gained tremendous attention in both academia and industry due to its inherent capability of representing arbitrary structures. Due to exponential increase in interest towards renewable sources of energy, especially wind energy, accurate wind speed forecasting has become very important. . In this paper, we propose a novel deep learning architecture, Multi Scale Graph Wavenet for wind speed forecasting. It is based on a graph convolutional neural network and captures both spatial and temporal relationships in multivariate time series weather data for wind speed forecasting. We especially took inspiration from dilated convolutions, skip connections and the inception network to capture temporal relationships and graph convolutional networks for capturing spatial relationships in the data. We conducted experiments on real wind speed data measured at different cities in Denmark and compared our results with the state-of-the-art baseline models. Our novel architecture outperformed the state-of-the-art methods on wind speed forecasting for multiple forecast horizons by 4-5%.

## I. Introduction

Weather forecasting is a very active area of research both for academia and the industry. Due to rapid climate change, occurrence of adverse weather events such as hurricanes, cyclones, floods and extreme rainfall has increased [1]. To counter human-induced climate change, governments across the world are shifting their focus to renewable sources of energy, especially wind and solar energy [2] and are using renewable energy forecasting to guide sustainable energy policies [3]. With increased adoption of wind energy as an alternative source, the number of wind energy parks has increased dramatically in recent years. To manage these wind energy parks efficiently and increase the contribution of wind energy in the energy mix, accurate wind speed forecasting for different time horizons is of paramount importance [4].

Traditionally, weather forecasting depends heavily on Numerical Weather Prediction (NWP) which uses mathematical formulations based on thermodynamics, fluid mechanics, etc. to simulate realistic weather conditions [5]. This requires huge amounts of computational resources and take several hours to forecast weather conditions [6]. For weather forecasting especially for less than three hours, NWP forecasting is not possible since these models take more than three hours for generating a forecast. Hence, there is a need to use alternative faster methods for short term weather forecasting.

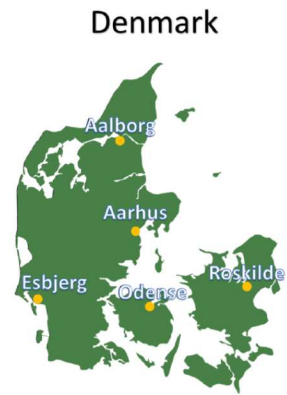

*Figure 1.Weather stations in Denmark [7]*

In recent years, deep learning has outperformed traditional methods in various tasks such as machine translation [8], speech recognition [9], image classification [10-11], image segmentation [12], etc. Deep learning-based architectures have been studied extensively for univariate [13] and multivariate time series regression [14], classification [15], anomaly detection [16] and remaining useful life estimation [17]. Very recently, graph convolutional neural networks (GCN) are being utilized by for irregular graph structured data like citation networks [18], social networks [19], molecular properties prediction for drug design [20], drug response [21], Quantum mechanical properties predictions for molecules [22], etc. These deep learning methods based on CNN and GCN are very fast and showed excellent performance in real time applications. These methods could efficiently address the problem for fast and accurate short term weather predictions.

NWP models for weather predictions are based on lot of assumptions about the underlying physical processes whereas in case of data-driven models, only historical data is utilized for weather forecasting [23]. Authors used 2D and 3D CNN to learn spatial and temporal relationship between different cities and variables. Reference [7] used multidimensional CNN for wind speed forecasting and outperformed the previous state-of-the-art on Denmark wind speed dataset [7]. Fig 1. shows the map corresponding to weather stations in Denmark for five cities namely Esbjerg, Aalborg, Aarhus, Odense and Roskilde. Many researchers tried combining convolutional neural network and graph convolutional network in the form of spatio-temporal graphs for traffic forecasting [24-26]. Recently, [27] used spatio-temporal graph convolutional neural network for wind speed forecasting along with learnable adjacency matrix that outperformed traditional machine learning models. Adjacency matrix with self-loop was also utilized to improve the forecasting accuracy of the models.

---
[1] represent equal contribution

The contribution of our work can be summarized as follows:

1) We propose a novel multi scale spatio-temporal GCN based on graph wavenet and inception network to forecast wind speed for multiple time horizons
2) We compare our results on benchmark wind speed forecasting dataset for Denmark and outperformed the state-of-the-art method for all four forecasting problems with different time horizons
3) We propose a single learnable adjacency matrix which explains the relative importance of neighboring nodes while forecasting the target variable for a given node.

The outline for rest of the paper is as follows. Section II describes the mathematical notations and baseline methods. Section III describes our novel multi scale GCN method. In Section IV, we discuss experimental details of our study. Lastly, in Section V, we compare our results with current state-of-the-art methods and present our conclusions.

## II. BACKGROUND

In this section, we mathematically summarize various machine learning approaches for spatio-temporal time series forecasting. Following notations are used in this paper to explain relevant concepts.

Let, Graph $G(V, E)$ represents a graph with vertices $V \in \{v_1, v_2, v_3 \ldots v_N\}$ and edges $E \in \{e_1, e_2, e_3, \ldots e_P\}$. Also, $X^t \in R^{NXD}$ is the input to the Graph at time $t$, where $N$ and $D$ are the number of nodes or vertices and number of features respectively. Given, $X^{(t-W):t} \in R^{DXNXW}$, our aim is to learn a function $f: [X^{(t-W):t}, G] \to Y^{t+T}$ where $W$ represents the time window size and T represents the future time instance for which prediction is required. The adjacency matrix corresponding to the graph $G(V, E)$ is $A \in R^{NXN}$. It signifies the strength of connection between different nodes of the graph and its value lies between 0 and 1. Small value of $A_{ij}$ signifies weak connection between node $i$ and node $j$. Similarly, large value of $A_{ij}$ signifies strong connection between node $i$ and node $j$. Note that if historical values of features corresponding to different nodes in a graph affect future values of features at those same nodes, we can use the following modification in the adjacency matrix.

$$\tilde{A} = A + \alpha I \quad (1)$$

Here, $\alpha$ is a learnable hyperparameter. Normalization of adjacency matrix is crucial as the strength between the nodes varies between 0 and 1. Let $\tilde{A}$ be the normalized adjacency matrix. We first use learnable node embeddings $E_1, E_2 \in R^{NXC}$ where $C$ is an integer hyperparameter. We propose the normalized adjacency matrix as

$$\tilde{A} = Softmax(E_1 E_2^T) \quad (2)$$

In eq. (2), softmax is applied row-wise over the matrix product $E_1 E_2^T$ so that each value in $i^{th}$ row denotes the relative strength of $i^{th}$ node with all other nodes in the graph.

### A. Wavenet Architecture

Here, we discuss the wavenet architecture proposed by [28]. Authors proposed wavenet architecture based on dilated convolutional layers for audio generation. These stacked dilated convolutional layers are explained below in Fig 2.

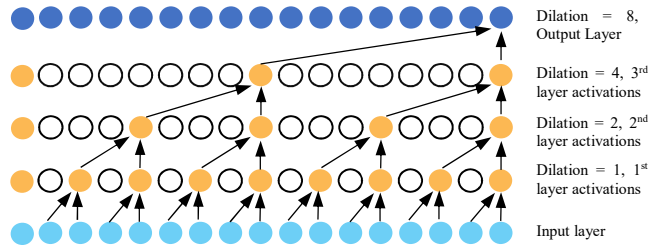

*Figure 2. Stacked Dilated Convolutional Layers for Time Series*

The receptive field can be increased by using stacked dilated convolutional layers while maintaining the same number of parameters as that of standard convolutional layers. Wavenet architecture consists of multiple stacked dilated convolutional layers and these dilated convolutional layers are connected through residual connections in a sequential manner. These residual connections improve the flow of gradient in the deep network containing many dilated convolutional layers. Also, the output of each of these dilated convolutional layers are parallelly connected through skip connections to capture features at multiple resolution. Reference [26] used spatio-temporal graph wavenet for traffic prediction. Authors used a linear layer to transform the input and then used gated temporal convolutional module (Gated-TCN) followed by graph convolutional module (GCN). Taking inspiration from the wavenet architecture, the spatio-temporal layers were connected through residual connections and the output of each spatio-temporal layer was skip connected to output layers.

In our study, we proposed two wavenet-inspired architectures for multivariate time series. The first architecture is similar to the one proposed by [26] and consists of several spatio-temporal blocks. In each spatio-temporal block a temporal block is followed by a graph convolutional layer. The second architecture consists of multi scale feature assimilation in each spatio-temporal block. We used both of our proposed architectures for wind speed forecasting at different time horizons and compared our results.

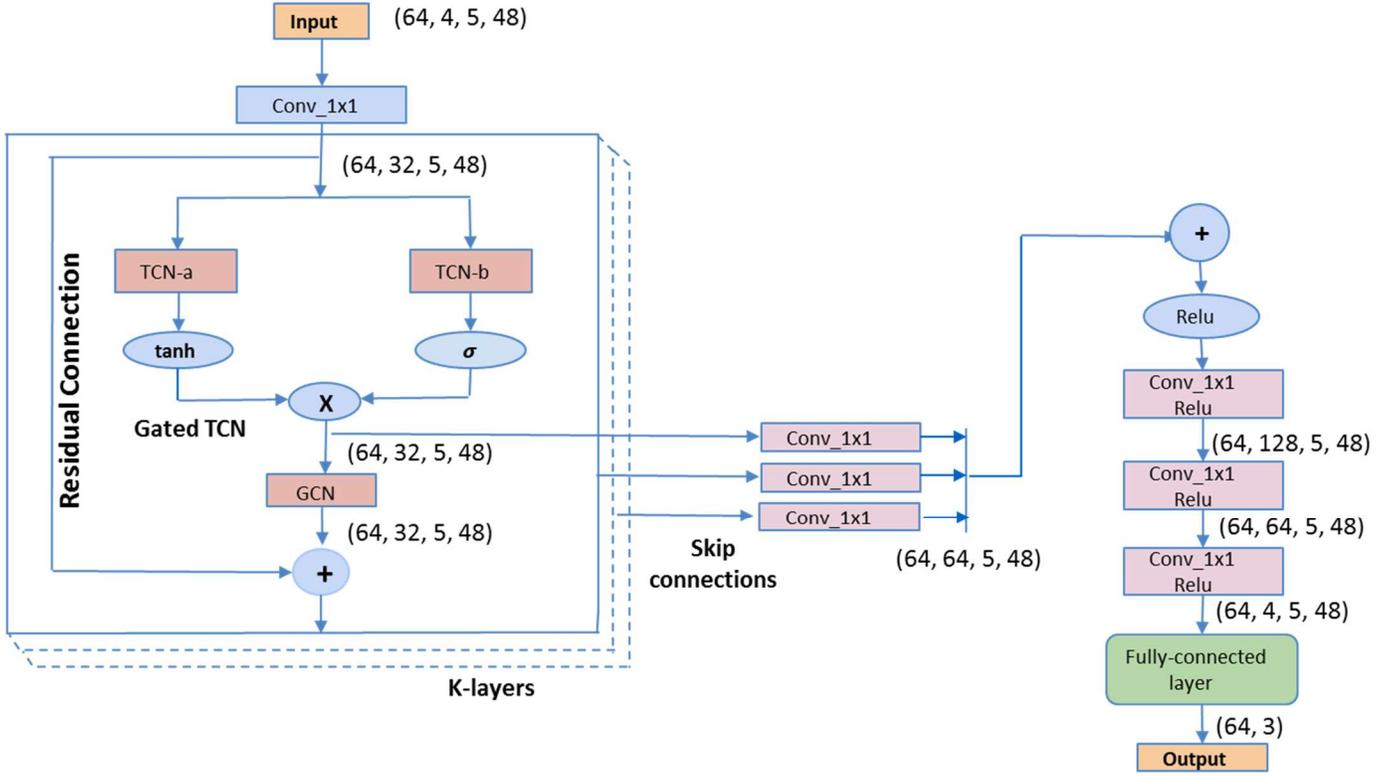

*Figure 3. Graph Wavenet for Spatio-temporal time series*

### III. PROPOSED NOVEL ARCHITECTURE

In this section, we describe our novel architecture based on spatio-temporal graph convolutional network and wavenet to solve the problem of multivariate time series forecasting across different nodes. Fig. 3 shows the overview of the proposed Graph wavenet architecture. We represent batch size, number of features, number of cities or nodes and the time window length as B, D, N and W respectively. The input dimension to Multi Scale Graph Wavenet is [B, D, N, W]. We first apply 1x1 convolutions to learn abstract local features and sequentially process the output through many spatio-temporal blocks. Spatio-temporal blocks are used to learn both spatial as well as temporal features. Each spatio-temporal block consists of Gated temporal convolutional network (Gated-TCN) unit and a Graph convolutional network (GCN) unit. We connect the output of each of the spatio-temporal block with the corresponding input using residual connection. We also apply 1x1 convolution to the output of each of the Gated-TCN unit and then add such outputs from all Gated-TCN units through skip connections. Subsequently, the sum of all the skip connections is passed through Rectified Linear Unit (RELU) and several 1x1 convolutional layers. Finally, the output of the last 1x1 convolutional layer is flattened and a fully connected layer (dense layer) is used to forecast the target variable at the target nodes. Now, we discuss the Multi Scale Graph Wavenet architecture. The TCN subunit of the same is shown in Fig 4. Kindly note that the same TCN subunit architecture is used in TCN-a and TCN-b subunits of the Gated-TCN unit. Also, note that architectural difference between Graph wavenet and Multi Scale Graph Wavenet lies only in the TCN subunit. Tanh and sigmoid activations are applied on the output of TCN-a and TCN-b respectively as used by [28]. In Multi Scale Graph Wavenet, we use filters of multiple sizes with various dilation rates and concatenate the output from all the filters before applying 1x1 convolutions. Using filters of different sizes with different dilations helps the model to efficiently collect information from a large receptive field and learn more robust features. We took the same adjacency matrix in all the GCN layers to make our model interpretable.

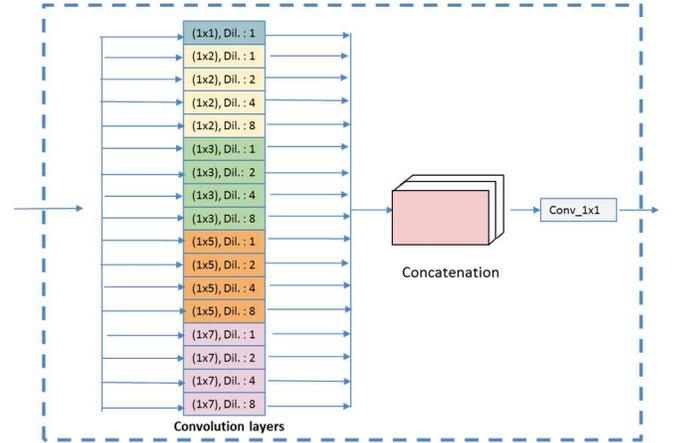

*Figure 4. TCN subunit for Multi Scale Graph Wavenet*

| Model | Mean Absolute Error (MAE) for | | | | Mean Squared Error (MSE) for | | | |
|---|---|---|---|---|---|---|---|---|
| | 6hr ahead forecast | 12hr ahead forecast | 18hr ahead forecast | 24hr ahead forecast | 6hr ahead forecast | 12hr ahead forecast | 18hr ahead forecast | 24hr ahead forecast |
| 2D | 1.304 | 1.746 | 1.93 | 2.004 | 2.824 | 5.088 | 6.12 | 6.61 |
| 2D + Attention | 1.313 | 1.715 | 1.905 | 1.95 | 2.885 | 4.896 | 5.933 | 6.201 |
| 2D + Upscaling | 1.307 | 1.723 | 1.858 | 1.985 | 2.826 | 4.931 | 5.639 | 6.474 |
| 3D | 1.311 | 1.677 | 1.908 | 1.957 | 2.855 | 4.595 | 5.958 | 6.238 |
| Multidimensional | 1.302 | 1.706 | 1.873 | 1.925 | 2.804 | 4.779 | 5.773 | 6.066 |
| WeatherGCNet | 1.279 | 1.638 | 1.777 | 1.869 | 2.698 | 4.407 | 5.148 | 5.641 |
| WeatherGCNet with $\gamma$ | 1.267 | 1.616 | 1.767 | 1.853 | 2.684 | 4.285 | 5.096 | 5.566 |
| **Graph Wavenet** | **1.261** | **1.597** | **1.794** | **1.876** | **2.626** | **4.119** | **5.052** | **5.564** |
| **Multi Scale Graph Wavenet** | **1.243** | **1.587** | **1.753** | **1.829** | **2.563** | **4.08** | **4.839** | **5.282** |
| % Improvement over WeatherGCNet with $\gamma$ | 1.89% | 1.79% | 0.79% | 1.30% | 4.51% | 4.78% | 5.04% | 5.10% |

*Table 1: Comparison of performance of wind speed forecasting for cities of Denmark for different time horizons*

## IV. EXPERIMENTAL DETAILS

In our study, we used Denmark wind speed forecasting data used by [27]. This dataset is collected from weather stations located in various cities of Denmark. The dataset contains variables such as ambient temperature, wind speed, wind direction and air pressure measured at hourly intervals. The data is recorded for 11 years starting from 2000 to 2010. We used 1 year of data each for validation and testing, and 9 years of data for training. Wind speed data corresponding to 2010 is used as the test data. Our target is to forecast wind speed in three cities namely, Esbjerg, Odense, and Roskilde for 6, 12, 18 and 24 hours in advance. We first arranged the data into 4 dimensional tensors of size [B, D, N, W] with B, D, N and W being 64, 4, 5 and 48 respectively. We used min-max normalization to normalize training, validation and test data. We evaluated our models using mean absolute error (MAE) and mean squared error (MSE) over the test dataset. We compared our results with the baseline models mentioned in [27] using the same test data and found that our models performed better for forecasting over different time horizons.

To train the network, we used Adam optimizer with a learning rate of 0.001. Learning rate is reduced by factor of 0.7 if the mean squared error loss over validation dataset does not improve for 3 consecutive epochs. We reload the model having the best performance on the validation dataset every time the learning rate is reduced. For each experiment, we trained the model for 50 epochs. We took 4 spatio-temporal blocks of TCN-GCN in both Graph Wavenet and Multi Scale Graph Wavenet.

## V. RESULTS AND DISCUSSION

In this section, we discuss and compare our results with existing baseline methods. The detailed comparison of our results with different methods is shown in Table 1. We used two methods namely, Graph Wavenet and Multi Scale Graph Wavenet. Graph Wavenet shows minor improvement as compared to the best baseline method, WeatherGCNet with $\gamma$ for 6, 12, 18 and 24 hours forecasting in terms of MSE. It performed better for 6 and 12 hours forecasting and slightly worse for 18 and 24 hours forecasting in terms of MAE. The Multi Scale Graph Wavenet method outperformed the best baseline models for all the 4 time horizon forecasting, both in terms of MAE and MSE.

The improvement in accuracy of the Multi Scale Graph Wavenet model can be attributed to filters of different sizes with different dilation rates in the Temporal Convolutional Network (TCN) subunit. The different filters in TCN subunit are able to learn temporal correlations at multiple time-scales. Filters with high dilation rate in the TCN subunit at different spatio-temporal blocks increase the effective receptive field size. Also, presence of filters of different dilation rates and sizes ensure that the model accumulates contextual information from multiple temporal resolutions for accurate forecasting. The residual connections between consecutive spatio-temporal blocks assist in better flow of gradients in all the spatio-temporal blocks. Improved performance of our model can also be attributed to efficient mixing of features at different time scales via skip connections from different spatio-temporal blocks. This helps the model to forecast more accurately at different time horizons. By treating different cities as nodes in a graph, we are able to learn spatial relationships among various nodes. The adjacency matrix is made learnable while training our model so that the spatial relationships between multivariate data of any two cities can be learned in an automated manner. Due to robust temporal and spatial features, Multi Scale Graph Wavenet is able to outperform all other models by a margin of 4-5% on MSE.

In our experiments, we used the same learnable adjacency matrix in all the layers in Graph Wavenet as well as Multi Scale Graph Wavenet. The learnt adjacency matrices for the 4 models of Multi Scale Graph Wavenet corresponding to 4 different forecast horizons are shown in Fig. 5. Each entry $A_{ij}$ of the adjacency matrix $A$ represents the relative importance of the $j^{th}$ node in improving weather forecasting for the $i^{th}$ node. Each row of the adjacency matrix is normalized i.e., sum of all the elements in each row is equal to 1. The learnt adjacency matrices seem to be consistent with geographical locations of different cities. For example, as shown in Fig. 5 weather forecasting for 6 hours horizon at City Odense is most influenced by weather conditions at Aarhus represented by the highest value of $A_{42}$ among $A_{4j}$, $j \in [1:5]$ shown in Fig. 5. Similarly, the adjacency matrices for all other forecast horizons such as 12 hours, 18 hours, and 24 hours show that wind speed at Odense is most influenced by weather conditions at Aarhus. This observation is consistent with the fact that Aarhus is geographically closest to Odense among all the other considered cities shown in Fig 1. Similarly, the geographical relationships for Roskilde and Esbjerg are captured by $A_{5j}$ and $A_{3j}$, $j \in [1:5]$ entries of the adjacency matrices. Since Roskilde and Esbjerg are the farthest cities (Fig. 1.), the effect of one on the other in the forecasting models at different horizons should be negligible. This can be verified by the obtained values in the adjacency matrices ($A_{53}$ and $A_{35}$) for different forecasting models shown in Fig. 5. Also, note that adjacency matrices are not symmetrical in general. Since wind speed is directional, the influence of one city on another city is not the same in the reverse direction. For example, $A_{34}$ denotes influence of Odense on Esbjerg and $A_{43}$ denotes influence of Esbjerg on Odense. $A_{34}$ and $A_{43}$ are significantly different denoting asymmetric influence.

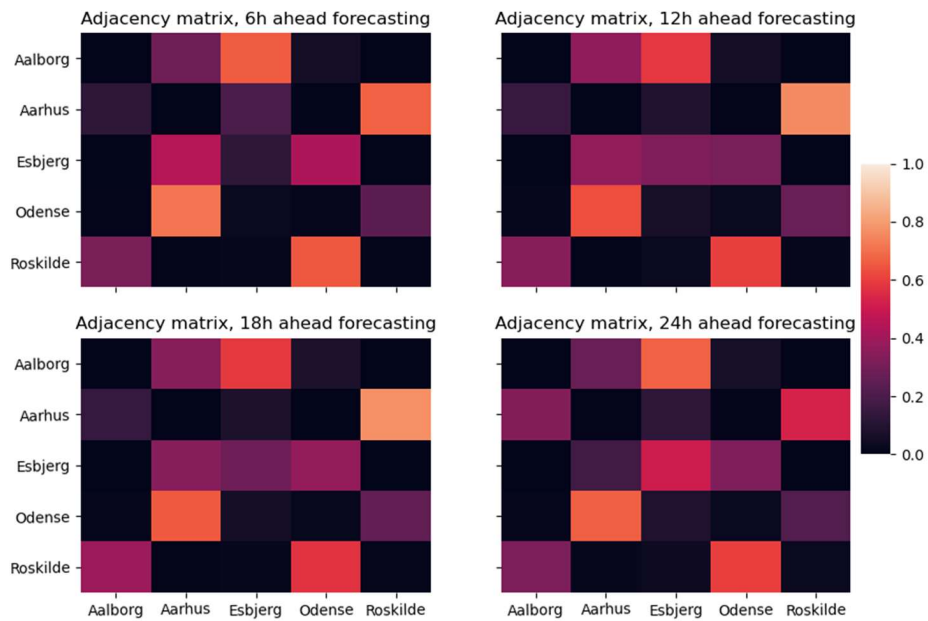

Figure 5. Visualization of the learnt adjacency matrices for different models

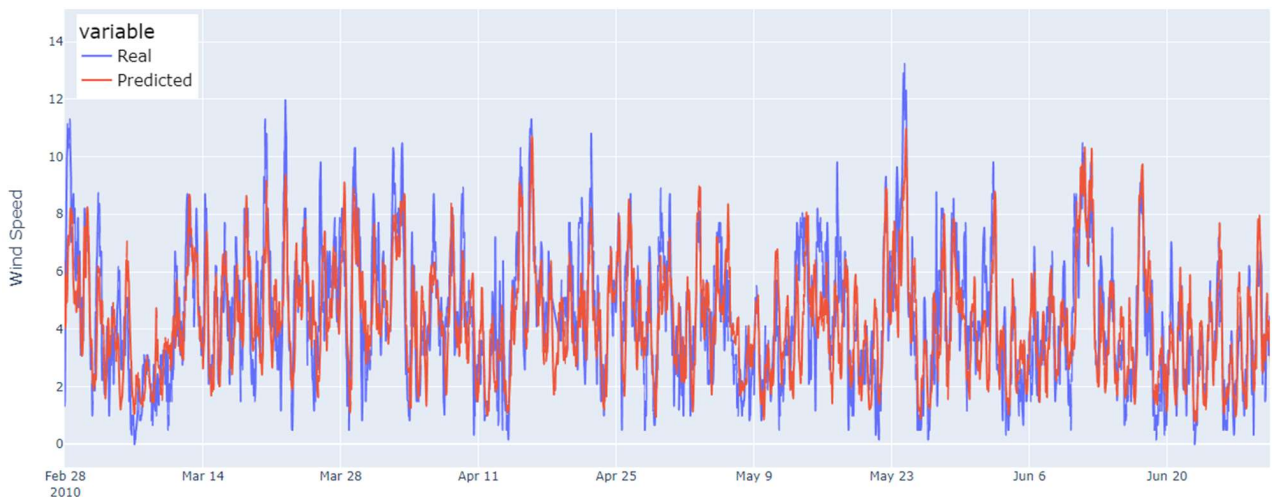

Figure 6. Time series plot of 6hr ahead real and predicted values of wind speed

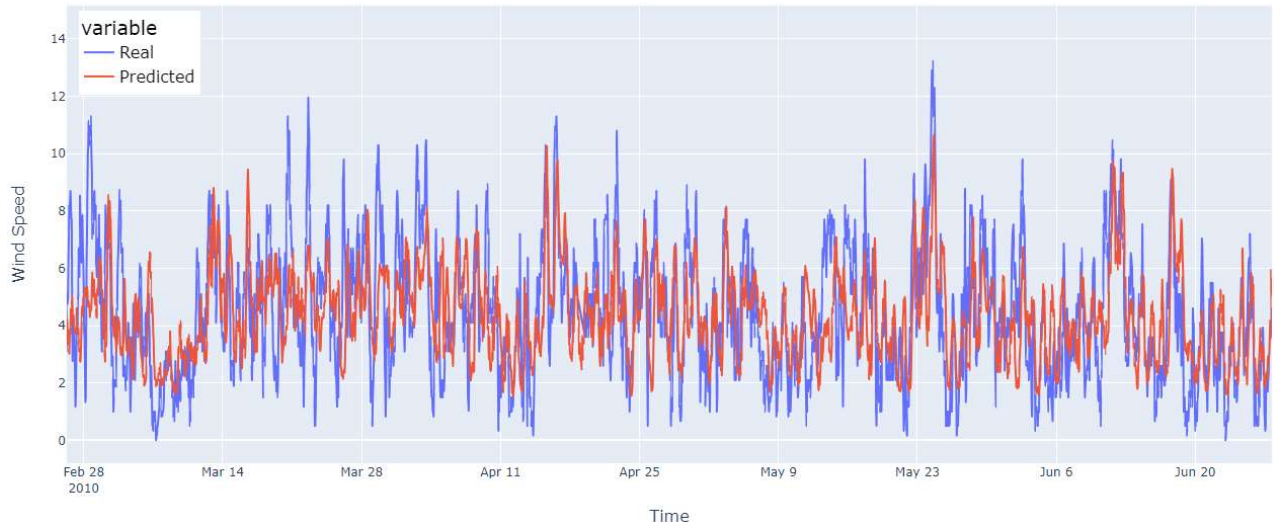

Figure 7. Time series plot of 12hr ahead real and predicted values of wind speed

We show the time series plots of real and forecasted wind speeds 6h and 12h ahead for Roskilde for a period of 4 months in Fig. 6 and Fig. 7 respectively. It can be observed from the figures that our models are able to capture the overall trends of change in wind speed quite well. However, MAE for the 12hr ahead forecast is higher than that for the 6h ahead forecast as longer horizon forecasting is more difficult than short horizon forecasting.

*A. Conclusion*

We propose a novel deep learning architecture, Multi Scale Graph Wavenet to capture spatio-temporal relationships among different cities to be used for wind speed forecasting. We demonstrated the effectiveness of our architecture and compared it with state-of-the-art methods available for wind speed forecasting for multiple cities in Denmark. We used filters of multiple sizes with multiple dilation rates to learn better features. Our models outperformed the state-of-the-art methods with 4-5% improvement in MSE. We also emphasize the role of a learnable adjacency matrix and showed that values of learnt adjacency matrices are consistent with the geographical positions of different cities.

ACKNOWLEDGMENT

We would like to thank the authors of "Deep Graph Convolutional Networks for Wind Speed Prediction" for providing their codes and datasets. We also thank the authors of "Graph WaveNet for Deep Spatial-Temporal Graph Modeling" for providing codes.